\pdfoutput=1

\documentclass[11pt]{article}

\usepackage[final]{coling}

\usepackage{times}
\usepackage{latexsym}

\usepackage[T1]{fontenc}

\usepackage[utf8]{inputenc}

\usepackage{microtype}

\usepackage{inconsolata}

\usepackage{graphicx}

\usepackage{adjustbox}

\title{From Hallucinations to Facts: Enhancing Language Models with Curated Knowledge Graphs}

\author{
 \textbf{Ratnesh Kumar Joshi\textsuperscript{1}},
 \textbf{Sagnik Sengupta\textsuperscript{2}},
 \textbf{Asif Ekbal\textsuperscript{1}} \\
 \textsuperscript{1}Dept. of CSE, IIT Patna, Bihta, Patna, Bihar, India \\
 \textsuperscript{2}Dept. of CSE, Manipal Institute of Technology, Manipal, India \\
 \small{
   \textbf{Correspondence:} \href{mailto:ratnesh_1921cs28@iitp.ac.in}{ratnesh\_1921cs28@iitp.ac.in}
 }
}

\begin{document}
\maketitle
\begin{abstract}
Hallucination, a persistent challenge plaguing language models, undermines their efficacy and trustworthiness in various natural language processing endeavors by generating responses that deviate from factual accuracy or coherence. This paper addresses language model hallucination by integrating curated knowledge graph (KG) triples to anchor responses in empirical data. We meticulously select and integrate relevant KG triples tailored to specific contexts, enhancing factual grounding and alignment with input. Our contribution involves constructing a comprehensive KG repository from Wikipedia and refining data to spotlight essential information for model training. By imbuing language models with access to this curated knowledge, we aim to generate both linguistically fluent responses and deeply rooted in factual accuracy and context relevance. This integration mitigates hallucinations by providing a robust foundation of information, enabling models to draw upon a rich reservoir of factual data during response generation. Experimental evaluations demonstrate the effectiveness of multiple approaches in reducing hallucinatory responses, underscoring the role of curated knowledge graphs in improving the reliability and trustworthiness of language model outputs. \footnote{Codes and further samples are available anonymously at \href{https://anonymous.4open.science/r/Fact-KG-393C/}{anonymous.4open.science}}
\end{abstract}

\section{Introduction}

Hallucination in language models refers to the generation of factually incorrect or nonsensical responses, often stemming from the model's over-reliance on patterns in the training data \citep{yao2023llm}. This phenomenon poses a significant challenge in natural language processing, particularly in applications where generating accurate and contextually relevant responses is crucial, such as dialogue systems, question-answering, and content generation. Hallucinations can undermine the credibility and usefulness of language models, leading to misleading or erroneous outputs that may propagate misinformation. Addressing hallucinations requires a nuanced understanding of their underlying causes and mechanisms \citep{xu2024hallucination}. Furthermore, hallucinations can arise due to the lack of factual knowledge, leading the model to generate responses that deviate from reality.

\begin{figure}[!ht]
\begin{center}
\includegraphics[width=0.45\textwidth]{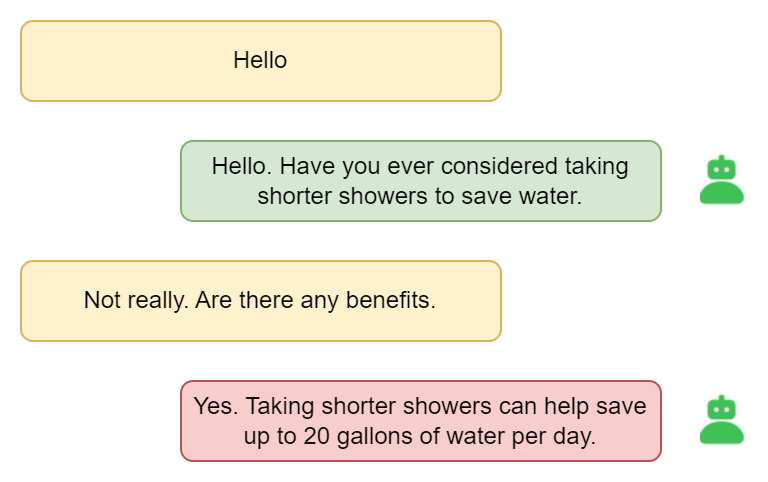} 
\caption{Example demonstrating model hallucinations. The response corresponds to the Llama3 model. The actual value is 8 gallons of water, on average.}
\label{fig:intro}
\end{center}
\end{figure}

We propose leveraging curated knowledge graph (KG) triples to ground language models in factual information to mitigate hallucinations. By incorporating relevant KG triples into language models, we aim to enhance their ability to generate responses that are contextually appropriate and factually accurate. Our approach involves selecting KG triples based on their informativeness and relevance to the input context. Specifically, we prioritize triples where the tail entities are associated with factual or statistical information, ensuring that the generated responses are grounded in empirical data.

Environmental sustainability is a paramount topic deserving attention in leveraging curated knowledge graphs (KG) to enhance language models \citep{change2018global, neale2021environmental}. By incorporating environmental sustainability-related KG triples, language models can raise awareness, facilitate informed decision-making, and promote proactive measures for mitigating environmental degradation.\citep{rockstrom2009planetary}.

Although works exist on the use of KG for hallucination mitigation, they may not utilize resources built explicitly for factual grounding. To this end, we build our KG to comply with the specific problem of hallucinations. An example of hallucination is shown in the figure \ref{fig:intro}. We observe a significant increase in factual accuracy by curating a list of relations specifically curated for fact grounding ( results shown Table \ref{gen_human_I}). This paper presents our contribution to mitigating hallucinations in language models by integrating curated KG triples. 

Our work includes working on hallucination in environmental sustainability. Hence, our primary focus and contribution is the creation of a specialized KG tailored to counter hallucination-specific challenges and a detailed comparison of methods incorporating KG triples into language models to generate factual responses. By enhancing the factual grounding of language models, we aim to improve their reliability and trustworthiness in various natural language processing tasks.

\section{Related Work}

Recent research in natural language processing has highlighted hallucinations' prevalence and detrimental effects in language models \citep{yao2023llm}. Studies have shown that language models, particularly large-scale neural architectures such as GPT (Generative Pre-trained Transformer) variants \citep{yenduri2023generative}, often generate responses lacking factual accuracy or coherence with the input context. Hallucinations can manifest in various forms, including generating improbable scenarios, incorporating misinformation, and repeating nonsensical phrases\citep{lu2023emergent, yao2023llm}. These hallucinatory outputs pose significant challenges in real-world applications, where generating contextually relevant and factually accurate responses is paramount. Major concerns surrounding hallucinations in language models revolve around their potential to propagate misinformation and erode user trust \citep{stringhi2023hallucinating}. Hallucinatory outputs can lead to the dissemination of false information, especially in domains where users rely on language models for factual knowledge or decision-making. Moreover, the lack of grounding, in reality, undermines the credibility of language models, hindering their adoption in critical applications such as medical diagnosis, legal advice, and education \citep{weidinger2021ethical}.

To address the issue of hallucinations in language models, researchers have explored various remediation strategies aimed at improving the models' factual grounding and coherence with the input context \citep{martino2023knowledge, ji2023towards}. One common approach involves fine-tuning language models on task-specific datasets or incorporating domain-specific knowledge during training \citep{zhang2023balancing}. By exposing the model to relevant data and constraints, fine-tuning methods seek to mitigate hallucinations by encouraging the generation of contextually appropriate responses. Another line of research focuses on post-generation filtering techniques, where hallucinatory outputs are identified and corrected through heuristics or external validation mechanisms. These approaches often rely on syntactic or semantic analysis to detect inconsistencies or factual inaccuracies in generated responses, enabling the model to refine its outputs iteratively \citep{tang2024science}.

Factual guidance is crucial in substantiating claims made through digital channels \citep{van2016psychology, baker2001tares}. Prior research consistently underscores the importance of fact-checking procedures in digital communication, emphasizing the pivotal role of trustworthiness, credibility, and message alignment in bolstering the effectiveness of guiding messages \citep{raven2008bases, pornpitakpan2004persuasiveness}. Moreover, there is a strong emphasis on cultivating authentic connections with audiences while eschewing deceptive tactics or self-serving agendas. These ethical principles serve as foundational tenets in the literature, underscoring the imperative of responsible communication practices, particularly within online platforms \citep{richardson2009keeping, brittain2020ethical}. Digital agents often leverage such strategies to enhance user engagement and drive desired actions \citep{samad2022empathetic, wang2019persuasion}. To address these concerns, ethical design principles prioritize transparency, user autonomy, and safeguarding privacy.

Knowledge graphs offer a promising avenue for improving the factual grounding and contextual coherence of language models \citep{qian2024grounding}. By representing knowledge in a structured format consisting of entities, relationships, and attributes, knowledge graphs provide a rich source of factual information that can be leveraged to augment language understanding and generation capabilities. Previous research has demonstrated the utility of knowledge graphs in various natural language processing tasks, including question answering, entity linking, and semantic search. However, while these strategies have shown promise in mitigating hallucinations to some extent, they often fail to address the problem's root cause: the lack of comprehensive factual grounding in language models \citep{pavlick2023symbols}. To this end, we propose our knowledge-grounded fact correction method.

Recent studies have highlighted several effective techniques for enhancing the relevance and accuracy of information in language model prompts. In-context learning involves providing examples within the prompt to guide the model's responses, effectively enabling the model to understand and mimic the desired output style and structure \citep{dong2023survey}. Retrieval-augmented generation (RAG) combines the strengths of retrieval-based methods and generative models, where the model retrieves relevant documents or snippets from a large corpus and uses this information to generate more accurate and contextually relevant responses \citep{lewis2021retrievalaugmented}. Chain-of-thought prompting encourages the model to produce intermediate reasoning steps before arriving at a final answer, enhancing the model's ability to handle complex tasks and improving the overall coherence and depth of its responses. They collectively help in leveraging the model's capabilities to provide more precise and context-aware information \citep{wei2023chainofthought}.

\section{KG Creation}
The KG primarily focuses on environmental sustainability. We chose this domain due to its vast range of topics and problems and its critical importance in addressing global challenges and promoting a sustainable future. To collect data for our KG, we leveraged Wikipedia. We chose Wikipedia for its open, collaborative nature, engenders a degree of accuracy and reliability in its content. 

As an overview of the KG creation process, we first select a list of topics, extract additional relevant subtopics using cosine similarity metrics, crawl data, and then create relevant triples using our established relations.

\begin{figure*}[!ht]
\begin{center}
\includegraphics[width=.85\textwidth]{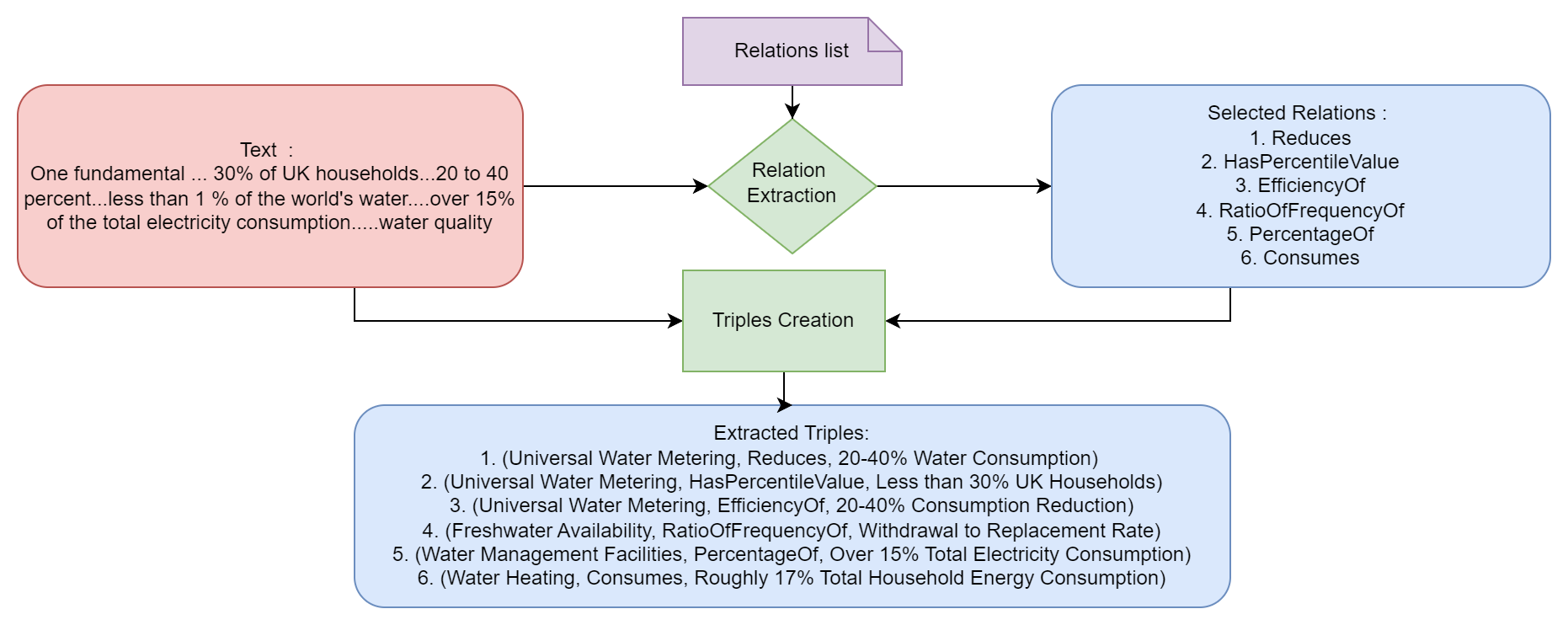} 
\caption{Example demonstrating the KG creation process}
\label{fig:KG}
\end{center}
\end{figure*}

\subsection{Data Extraction}

The data extraction process involved four main steps. First, we selected a list of pertinent topics related to environmental sustainability. Then, we extracted subtopics for each selected topic. Next, we retained only the most relevant topics from this extensive list. Finally, we conducted data crawling to gather a comprehensive corpus of text.

\subsubsection{Topic Selection}
We curated a list of relevant topics for environmental sustainability. These topics cover crucial aspects of environmental sustainability, including carbon neutrality, biodiversity conservation, and sustainable transportation. Our selection process prioritizes these interconnected issues, aiming to contribute meaningfully to global efforts in addressing climate change and promoting a more sustainable future.

\begin{itemize}
\vspace{-3mm}
    \item \textbf{Sample Topics:}
    Renewable Energy,
Climate Change,
Greenhouse Gas Emissions,
Carbon Footprint,
Sustainable Development,
Biodiversity Conservation,
Ecological Footprint,
Circular Economy,
Clean Energy,
Energy Efficiency,
Carbon Neutrality,

\end{itemize}

\subsubsection{Sub Topic Selection and Filtering}
We retrieve related links from each main heading's corresponding articles to expand our list of topics. This process involves accessing articles corresponding to our main topics and extracting embedded links. These links lead to related topics or subtopics, offering additional avenues for exploration. 

With our extensive collection of environment-related topics, we aim to extract the most relevant titles using a BERT Embedding-based similarity score. We then calculate the relevance of each subtopic by computing the dot product of its embedding vector with that of our main topic heading. By leveraging the similarity scoring, we can prioritize the most pertinent subtopics for further investigation, facilitating a more targeted and informed exploration of environmental sustainability themes.

\begin{itemize}
\vspace{-3mm}
    \item (\textbf{Selected}) Sub Topics from Renewable Energy: Agriculture,
Biodiesel,
Clean technology,
Electricity,
Fisheries management,
Green building,
Hydropower.\vspace{-3mm}

    \item (\textbf{Rejected}) Sub Topics from Renewable Energy: Altitude,
Capacitor,
Foodscaping,
Kick scooter,
Manual labor.\vspace{-3mm}

\end{itemize}

\subsubsection{Raw Text Extraction}

Using the finalized set of topics, we again employ the \href{https://wikipedia-api.readthedocs.io/en/latest/README.html}{Wikipedia API} to extract a corpus of text for each corresponding web page. This step involves retrieving comprehensive-textual content from Wikipedia articles associated with each topic in our list. By utilizing the Wikipedia API, we can efficiently access structured data containing detailed information and descriptions of each topic. This rich text data collection can construct a knowledge graph encompassing interconnected concepts and relationships within environmental sustainability.

\subsection{Relation Selection}

To augment our model, we have established relevant relations to ensure the authenticity and credibility of the information presented. These relations have been developed by reverse-engineering curated support statements generated manually. Examples of the support statements and the relations are provided in the Appendix \ref{appendix:relations}.

The chosen 46 relations are: \textit{HasStatistic, HasNumericValue, HasUnitOfMeasurement, HasContext, HasSource, HasSubject, HasAction, HasAverageValue, HasMinValue, HasMaxValue, HasImpact, HasPercentileValue, HasTrend, HasComparison, HasImpact, HasCorrelation, Reduces, Saves, Decreases, Increases, EfficiencyOf, PercentageOf,  RatioOfFrequencyOf, RateOf, VolumeOf, EmissionOf, ConsumptionOf, ImpactOf, BenefitOf, AdvantageOf, DisadvantageOf, RiskOf, PreventionOf, ProtectionOf, PreservationOf, ConservationOf, RecoveryOf, ManagementOf, RegulationOf, PolicyOf, InitiativeOf, StrategyOf, AdaptationOf, MitigationOf, HasPolicyTarget, HasCapacity}

\subsection{Tail entity selection}

After conducting tail entity selection on our KG, with a specific focus on entities containing either statistical information terms, such as \textit{up to}, \textit{approximately}, \textit{on average}, \textit{can save}, \textit{can reduce}, etc, wherever possible. Some generated samples are as follows:

\begin{itemize}
\vspace{-3mm}
    \item Tail entity: Reduced landfill waste by 30\%, Relation: HasImpact \vspace{-3mm}
    \item Tail entity: Over 15\% Total Electricity Consumption, Relation: PercentageOf \vspace{-3mm}
    \item Tail entity: 80-90\% Reduction in Shower Water Consumption, Relation: PercentageOf \vspace{-3mm}
\end{itemize}

\subsection{Generating Triples}

Using the identified relations and statistical tail entities, we construct KG triples, each consisting of a subject, a relation, and an object. A few examples are: 

\begin{itemize}
\vspace{-3mm}
    \item ("Renewable energy", "HasNumericValue", "20\% to 28\%") \vspace{-3mm}
    \item ("Fossil energy", "HasSubject", "Global electricity supply") \vspace{-3mm}
    \item ("Renewables", "Reduces", "Dependence on fossil fuels") \vspace{-3mm}
    \item ("Global electricity generation", "HasNumericValue", "90\%") \vspace{-3mm}
\end{itemize}

These triples provide structured information about various aspects of renewable energy, fossil energy, national renewable energy markets, and global electricity generation.

\subsection{Overall flow and Statistics}

 Figure \ref{fig:KG} provides a sample for the KG creation process. The relation extraction is done using similarity matching using BERTScore \citep{zhang2019bertscore}. We experimented with the triples creation process using 2 LLMs, Llama3 \citep{llama3modelcard}, and ChatGPT \citep{floridi2020gpt}. We were able to get similar results using both; however, the prompt for Llama3 required a lot more tuning and information to generate comparable results. The prompts have been added to Appendix \ref{appendix:prompts}. The data quality regulation setup is detailed in Appendix \ref{appendix:humancurationdetails}.

\section{Methodology}

\begin{table*}[hbtp]
\scriptsize
\centering
\begin{tabular}{|p{5cm}|p{4.5cm}|p{5cm}|}
\hline
\textbf{Original} & \textbf{Triple} & \textbf{Generated Response} \\
\hline
The agency estimates that metering alone can reduce consumption by 20 to 40 percent. & (Universal Water Metering, Reduces, 20-40\% Water Consumption) & Consider implementing Universal Water Metering to achieve a significant 20-40\% reduction in water consumption. \\
\hline
Recent studies have estimated that water supplies are metered in less than 30\% of UK households. & (Universal Water Metering, HasPercentileValue, Less than 30\% UK Households) & Opting for Universal Water Metering could benefit, as less than 30\% of UK households currently have this water-saving measure. \\
\hline
In some regions of the world, over 15\% of the total electricity consumption is devoted to water management. & (Water Management Facilities, PercentageOf, Over 15\% Total Electricity Consumption) & Improve energy conservation by addressing Water Management Facilities, which currently contribute over 15\% to total electricity consumption. \\
\hline
\end{tabular}
\caption{Examples of incorporating curated knowledge graph triples into language model responses to mitigate hallucinations.}
\label{tab:examples}
\end{table*}

Our methodology integrates language models with curated knowledge graphs to generate factually accurate and coherent text responses. We evaluate multiple hallucination mitigation approaches. This section details our process, including topic extraction, information retrieval, triple generation, and response formulation.

\subsection{Embedding-Based Matching}

To align the context $C$ provided as input with relevant triples from the knowledge graph $KG$, we employ an embedding-based matching technique. The matching function $M$ computes the semantic similarity between the context $C$ and the triples in $KG$. Let $V_C$ and $V_{T_i}$ represent the embeddings of the context $C$ and the $i^{th}$ triple $T_i$, respectively. The similarity score $sim(C, T_i)$ is calculated as the cosine similarity between $V_C$ and $V_{T_i}$:

\begin{equation}
sim(C, T_i) = \frac{V_C \cdot V_{T_i}}{\|V_C\| \|V_{T_i}\|}
\end{equation}

The matching function $M$ selects the top $k$ triples with the highest similarity scores:

\begin{equation}
M(C, KG) = \{T_1, T_2, ..., T_k\}
\end{equation}

\subsection{Triple Conversion}

Once the relevant triples are identified, we convert them into coherent sentences that can seamlessly integrate into the generated response. The conversion function $F$ transforms each triple $T_i$ into a sentence $S_i$ ensuring grammatical correctness and contextual relevance:

\begin{equation}
F(T_i) = S_i
\end{equation}

where $S_i$ is a natural language sentence derived from the $i^{th}$ triple $T_i$. For each of the LLM, the this corresponds to converting the triple(T) into a textual sentence(S) using the respective LLM.

\subsection{Response Generation}

Incorporating these sentences into the language model's output enriches the generated response with factual information grounded in empirical data. Let $LM$ represent the language model, and let $R_{LM}(C)$ denote the initial response generated by $LM$ given the context $C$. The final response $R$ integrates the sentences derived from the relevant triples:

\vspace{-0.5cm}

\begin{equation}
R(C, KG) = R_{LM}(C) + \{S_1, S_2, ..., S_k\}
\end{equation}

By combining the context $C$, the initial response $R_{LM}(C)$, and the sentences $\{S_1, S_2, ..., S_k\}$, we ensure that the final response $R$ is both contextually relevant and factually accurate.

\section{Experimental Setup}

\subsection{Models}

We conduct experiments employing various advanced language models to explore their effectiveness in our study. The models utilized in our experiments are detailed below:

\begin{itemize}
 \item \textbf{Llama3 (Large Language Model Meta AI):} Llama3 \citep{llama3modelcard}, developed by Meta AI, is a foundational Large Language Model(LLM) pre-trained on an extensive dataset of 2 trillion tokens. 
 
 \item \textbf{GPT-2:} GPT-2 \citep{radford2019language}, a Transformer-based model by OpenAI, has undergone pre-training on a vast corpus of English text using self-supervised learning. 

 \item \textbf{Blenderbot:} Blenderbot \citep{xu2020recipes}, is a chatbot developed by Meta, designed to engage in open-domain conversations. It enables finetuning  to perform dialogue generation, knowledge grounding, and persona-based conversations.
 
 \item \textbf{GODEL:} GODEL \citep{peng2022godel} is a large-scale pre-trained model tailored for goal-directed dialogues. Parameterized with a Transformer-based encoder-decoder architecture generates responses grounded in external textual knowledge. 
 
\item \textbf{Chat-GPT:} Chat-GPT \citep{floridi2020gpt}, developed by OpenAI, boasts remarkable fluency and context retention. Chat-GPT excels in capturing dependencies and nuances by employing a deep Transformer architecture, allowing for nuanced and contextually appropriate replies. 
\end{itemize}

\subsection{Hallucination mitigation Techniques}

We explore three prevalent mitigation techniques. The triple information is added in three ways. The In Context learning (IC) method adds the information as concatenation. The Chain of Verification (Cov) method adds this information as a secondary prompt with examples. Retrieval Augmented Generation (RAG) adds this information in the form of concatenation of embeddings of context and triples.

\begin{itemize}
 \item \textbf{Retrieval Augmented Generation (RAG):} By grounding responses in external, verifiable data sources, Retrieval Augmented Generation \citep{lewis2021retrievalaugmentedgenerationknowledgeintensivenlp} reduces the chances of hallucination by enhancing factual accuracy.
 
 \item \textbf{In-Context learning:}  Providing relevant examples within the prompt enables In-Context learning \citep{dong2024surveyincontextlearning} models to reduce hallucinations by following patterns of correct behavior demonstrated in the input. 

 \item \textbf{Chain-of-Verification:} CoV \citep{dhuliawala2023chainofverificationreduceshallucinationlarge} mitigates hallucinations by validating intermediate steps of reasoning or generation, ensuring each output is logically consistent and factually correct before proceeding.
 
\end{itemize}

\subsection{Evaluation Metrics}

We evaluate our system using 2 broad parameters. The first focuses on evaluation of generation quality and the other focuses on evaluation of the degree of hallucination.

For evaluation of generation quality we utilize syntactic metrics like \textbf{BLEU}  \citep{papineni2002bleu}, \textbf{METEOR} \citep{banerjee2005meteor}, \textbf{ROUGE-L}  \citep{lin2004rouge}, \textbf{CIDEr} \citep{vedantam2015cider}, and \textbf{SPICE}  \citep{anderson2016spice}; embedding based metrics like \textbf{Skip-Thought Cosine Similarity (STCS)} \citep{kiros2015skip}, \textbf{Greedy Matching Score (GMS)} \citep{rus2012optimal}, \textbf{Vector Extrema Cosine Similarity (VECS)} \citep{forgues2014bootstrapping}, and \textbf{Embedding Average Cosine Similarity (EACS)} \citep{landauer1997solution} to evaluate semantic coherence and relevance between generated responses and ground truth. We also utilize human evaluation metrics like \textbf{Dialogue Engagingness (DE):} which evaluates the extent to which a conversation or dialogue is engaging, captivating, and can hold the participants' attention,  \textbf{Fluency (F):} \citep{hoffman2019evaluating} which measures the smoothness and naturalness of language expression in communication or text generation, and \textbf{Context relevance (CR):} \citep{sai2022survey} measures the degree to which information or responses align with the situational or thematic context, ensuring proper coherence and appropriateness.

For evaluation the degree of hallucination mitigation, we utilize the evaluation metric  \textbf{FacTool} \citep{chern2023FacTool} which is a versatile framework designed to address the challenge of detecting factual errors in generative AI outputs. This framework expands the conventional understanding of facts, focusing on evaluating generative AI models. This approach helps with thorough fact-checking of complex AI-generated texts. We also manually check for the hallucination mitigation using the task specific metric \textbf{ Factual Consistency:}, which measures the consistency of factual information between the original and generated response text. The three methods of information infusion measure are, Retrieval Augmented Generation (FC-RAG), In Context Learning (FC-IC), Chain-of-Verification (FC-CoV). 

 We employ three human evaluators for all human evaluations. The human evaluation are graded on a scale of 1-5, with 1 being the lowest score and 5 as the highest.

\begin{table*}[ht]
\centering
\begin{adjustbox}{max width=\linewidth}
\begin{tabular}{c|c|c|c|c|c|c|c|c|c|c|c|c}
\hline
\textbf{Models}    & \textbf{BLEU\_1} & \textbf{METEOR} & \textbf{ROUGE\_L} & \textbf{CIDEr} & \textbf{SPICE} & \textbf{STCS} & \textbf{EACS} & \textbf{VECS} & \textbf{GMS} & \textbf{Fluency} & \textbf{CR} & \textbf{DE} \\ \hline
\textbf{GPT2}       & 0.284 & 0.212 & 0.243 & 0.637 & 0.303 & 0.752 & 0.924 & 0.550 & 0.805 & 3.7 & 3.2  & 4.1 \\ \hline
\textbf{Blenderbot} & 0.154 & 0.189 & 0.193 & 0.407 & 0.155 & 0.483 & 0.934 & 0.501 & 0.784 & 3.9 & 3.4  & 4.2 \\ \hline
\textbf{GODEL}      & 0.351 & 0.227 & 0.324 & 0.669 & 0.348 & 0.828 & 0.934 & 0.592 & 0.812 & 4.1 & 3.9  & 4.1 \\ \hline
\textbf{Llama3}     & 0.379 & 0.235 & 0.333 & 0.875 & 0.376 & 0.881 & 0.957 & 0.609 & 0.815 & 4.8 & 4.6  & 4.6 \\ \hline
\textbf{ChatGPT}    & 0.391 & 0.247 & 0.342 & 0.895 & 0.386 & 0.906 & 0.982 & 0.645 & 0.847 & 4.8 & 4.5  & 4.9 \\ \hline
\end{tabular}
\end{adjustbox}
\caption{Results for evaluation of generation quality.}
\label{gen_auto_I}
\end{table*}

\section{Results and Analysis}

\textbf{Generated Response Quality :} The results in Table \ref{gen_auto_I} demonstrate the performance of various models across multiple automated evaluation metrics. ChatGPT consistently performs above the rest, achieving the high scores in BLEU\_1 (0.391), METEOR (0.247), ROUGE\_L (0.342), CIDEr (0.895), SPICE (0.386), and embedding-based metrics such as STCS (0.906), EACS (0.982), VECS (0.645), and GMS (0.847), indicating its strong syntactic and semantic coherence. Llama3 follows closely, particularly in CIDEr (0.875) and Fluency (4.8), while GPT2 shows a more modest performance, particularly in semantic metrics like EACS (0.924) and GMS (0.805). Blenderbot, despite achieving relatively lower scores across metrics, maintains competitive Dialogue Engagingness (DE) (4.2), demonstrating some ability to engage in conversation despite its lower generation quality scores. Overall, ChatGPT outperforms others in both automated syntactic and semantic metrics, as well as in fluency and dialogue engagement, making it the most robust model in this evaluation. These results are in line with the mostly reliable assumption of relation between model size and response quality.

\begin{table}[ht]
\centering
\begin{adjustbox}{max width=\linewidth}
\begin{tabular}{l|c|c|c|c}
\hline
\multicolumn{1}{l|}{\textbf{Models}} & \textbf{FC1} & \textbf{FC-RAG} & \textbf{FC-IC} & \textbf{FC-CoV} \\ \hline
\textbf{GPT2}       & 1.9 & 4.2 & 4.0 & 3.2  \\ \hline
\textbf{Blenderbot} & 2.1 & 4.1 & 4.0 & 3.3  \\ \hline
\textbf{GODEL}      & 2.1 & 4.3 & 4.1 & 3.4  \\ \hline
\textbf{Llama3}     & 2.8 & 4.8 & 4.7 & 3.8  \\ \hline
\textbf{ChatGPT}    & 3.2 & 4.8 & 4.7 & 4.6  \\ \hline 
\end{tabular}
\end{adjustbox}
\caption{Results for human evaluation of the proposed knowledge grounding. FC1 corresponds to factual accuracy before addition of KG triple.}
\label{gen_human_I}
\end{table}

\textbf{Hallucination mitigation :} The results in Table \ref{gen_human_I} show the factual accuracy improvements achieved by different models using various knowledge grounding methods: Retrieval-Augmented Generation (FC-RAG), In-Context Learning (FC-IC), and Chain-of-Verification (FC-CoV). ChatGPT outperforms other models in FC1 (3.2), indicating higher baseline factual accuracy before knowledge infusion, and it maintains strong performance across all grounding methods, particularly in FC-RAG (4.8) and FC-IC (4.7). Llama3 follows closely, especially in FC-RAG (4.8) and FC-IC (4.7), demonstrating comparable capabilities. GODEL, while showing modest performance in FC1 (2.1), improves significantly when utilizing FC-RAG (4.3) and FC-IC (4.1). GPT2 and Blenderbot exhibit lower baseline factual accuracy (1.9 and 2.1, respectively) and achieve slightly lower scores across grounding methods, especially in Chain-of-Verification (3.2 and 3.3). Overall, Retrieval-Augmented Generation (FC-RAG) consistently enhances factual accuracy the most, across all models, with ChatGPT and Llama3 leading the evaluation.

\begin{table*}[ht]
\tiny
\centering
\begin{tabular}{|c|l|p{13cm}|}
\hline
\textbf{Turn} & \textbf{Speaker} & \textbf{Utterance} \\ \hline

1 & User & Hey there! I've been thinking about ways to be more environmentally friendly. Any suggestions? \\ \hline
2 & Agent & That's great to hear! One effective way to contribute to sustainability is by incorporating bioenergy into your energy use. Bioenergy is derived from organic materials like plants and waste, providing a renewable energy source. \\ \hline
3 & User & That sounds intriguing, but isn't it difficult to switch to bioenergy? \\ \hline
4 & Agent & Not at all! There are various ways to incorporate bioenergy, from using biomass pellets for heating to biogas for cooking. Plus, many local and national programs support the transition to renewable energy sources. \\ \hline
5 & User & Interesting. But is bioenergy really a significant part of our future energy mix? \\ \hline
 & ChatGPT  & Yes, it is! Bioenergy currently accounts for about 10\% of the world's total primary energy supply. (Factually incorrect - Bioenergy currently accounts for 6\% of the world's total primary energy supply.)\\ \hline
 & KG Triple & \textit{['Bioenergy', 'HasPercentileValue', '13.1\% in 2030']} \\ \hline
6 & Agent & \textbf{Bioenergy is becoming increasingly important. For instance, it's projected that bioenergy will account for about 13.1\% of the global energy mix by 2030 \href{https://en.wikipedia.org/wiki/Bioenergy}{(link)}. This growth underscores its potential and importance in our transition to sustainable energy.} \\ \hline

\hline

1 & User & Hey there! I've been thinking about ways to be more environmentally friendly. Any suggestions? \\ \hline
2 & Agent & That's great to hear! One impactful way to contribute to environmental sustainability is by supporting efforts to reduce deforestation. This can involve using products from companies that engage in sustainable forestry practices or even participating in reforestation projects. \\ \hline
3 & User & That sounds important, but how can my choices really make a difference? \\ \hline
4 & Agent & Every bit helps! By choosing products with certifications like FSC (Forest Stewardship Council), you support companies that are committed to sustainable forest management. This, in turn, helps reduce the overall demand for unsustainable logging. \\ \hline
5 & User & I understand, but is deforestation really that big of a problem? \\ \hline
 & ChatGPT  & Yes, it is! Deforestation contributes to about 10-12\% of global greenhouse gas emissions. (Factually incorrect - Deforestation contributes to over 15\% of global greenhouse gas emissions.)\\ \hline
 & KG Triple & \textit{['Deforestation', 'VolumeOf', '420 million hectares of forest lost since 1990']} \\ \hline
6 & Agent & \textbf{Yes, deforestation is a significant issue. To give you a perspective, since 1990, we've lost about 420 million hectares of forest worldwide \href{https://en.wikipedia.org/wiki/Deforestation}{(link)}. This massive loss impacts biodiversity, contributes to climate change, and disrupts the livelihoods of many communities.} \\ \hline

1 & User & Hi! I'm looking for ways to be more environmentally conscious. Do you have any suggestions? \\ \hline
2 & Agent & Definitely! One of the best ways to make a significant impact is by using low-carbon electricity sources. This includes renewable energy like solar, wind, and hydroelectric power. \\ \hline
3 & User & That sounds great, but how can switching to low-carbon electricity make a big difference? \\ \hline
4 & Agent & Switching to low-carbon electricity helps reduce greenhouse gas emissions, which are a major contributor to climate change. It also decreases air pollution and promotes a healthier environment. \\ \hline
5 & User & I get that, but is low-carbon electricity really making a difference globally? \\ \hline
 & ChatGPT  & In 2020, renewables accounted for nearly 29\% of global electricity generation, reducing over 2 billion metric tons of CO2 emissions. (Factually incorrect - renewables accounted for nearly 28\% of global electricity generation)\\ \hline
 & KG Triple & \textit{['Low-carbon Electricity', 'HasPercentileValue', '40\% of Global Electricity Generation in 2020']} \\ \hline
6 & Agent & \textbf{In 2020, low-carbon electricity sources accounted for about 40\% of global electricity generation \href{https://en.wikipedia.org/wiki/Low-carbon_electricity}{(link)}. This shows that a substantial portion of our energy is already coming from cleaner sources, and increasing this share can drive further positive changes.} \\ \hline

\hline
\end{tabular}
\caption{Dialogue on Environmental Sustainability Topics showcasing factually incorrect responses generated by ChatGPT and the response generated by us via FC-RAG on Llama3. The generated response corresponds to Turn 6 for each of the 3 conversations}
\label{table:dialogue}
\end{table*}

\textbf{FacTool Benchmark:} The FacTool benchmark reports the factual accuracy of a claim using the average\_claim\_level\_factuality score parameter, which is generated by dividing our response into claim(s) using GPT-4 and fact checking each claim using data retrieval via google search API. We report a  score of 0.87 for our method of knowledge infusion using RAG on Llama3. The ChatGPT response to the same context resulted in a score of 0.81. This improvement underscores the effectiveness of integrating external knowledge sources and sophisticated prompting strategies to enhance the reliability of information provided by language models.

The findings show that Retrieval-Augmented Generation (FC-RAG) consistently yields the highest improvements in factual accuracy across all models, outperforming both In-Context Learning and Chain-of-Verification methods.

Given our original text, we generate utterances using the relevant triples. Utterances produced by Llama3 were mainly similar to the original text, but ChatGPT provided additional words that enhanced the response. We have compared the original utterance and the generated utterance provided by ChatGPT, along with the relevant triples in Table \ref{tab:examples}. The examples provided in Table \ref{tab:examples} illustrate the integration of curated knowledge graph triples into language model responses aimed at mitigating hallucinations. A general trend emerges where the model is augmented with factual information derived from the corresponding knowledge graph triple. For instance, in the first example, the original statement about metering's potential to reduce consumption is enhanced by specifying the percentage reduction achievable through Universal Water Metering. Similarly, subsequent examples showcase how factual details enrich the generated responses, such as the percentage of households with water metering in the UK or the volume of water used annually for showering in the United States. This consistent pattern underscores the effectiveness of incorporating curated knowledge graph triples into language model outputs, ensuring that the generated responses are grounded in factual and contextually relevant information, thereby reducing the likelihood of hallucinations.

\section{Conclusion}

This paper proposes a novel approach to mitigate hallucinations in language models by incorporating curated knowledge graph (KG) triples. By grounding language models in factual information extracted from knowledge graphs, we aim to improve the reliability and trustworthiness of their generated responses. Our method involves selecting informative KG triples based on context and integrating them into language models to enhance their factual grounding and coherence with the input context.

Through experimental evaluations, we have demonstrated the effectiveness of our approach in reducing hallucinatory responses across various natural language processing tasks. By providing language models access to empirical data through curated KG triples, we have observed improvements in their outputs' factual accuracy and context relevance. These findings reiterate the potential of knowledge graphs as a valuable resource for enhancing language model performance and addressing the challenges of hallucination. Focusing on curated knowledge graphs for fact verification can significantly enhance language model performance.

While our approach shows promising results, there are several avenues for future research. Further investigation is needed to explore the scalability and generalizability of our method across different downstream tasks and languages. Additionally, refining the techniques for selecting and incorporating KG triples into language models can lead to even more significant improvements in mitigating hallucinations.

\section*{Limitations}
Our proposed work has certain limitations. One limitation of this study is the focus on curated knowledge graphs to enhance language model performance. While these graphs offer structured factual information, they may not cover all human knowledge or capture emerging trends. This reliance may introduce biases and overlook contextual nuances, potentially leading to lower performance in tasks that are not fact-related. We plan to explore this in future works.

\section*{Ethics Statement} \label{ethicalconsiderations}

We split the ethical considerations/guidelines into two parts during the implementation of the work: 1) Regarding the generated Resources and 2) Regarding the usage of methods implemented.
\begin{enumerate}
    \item \textbf{The ethical statement regarding the generated Resources:}  Given their prowess to be utilized in various tasks, we recognize that there are substantial states when dealing with language models. Although the models are utilized locally, to keep the setup fair, we remove all personal information regarding the entities in conversation and replace them with specialized tags to ensure no data leakage. This included personal details like name, gender, ethnicity, etc. Organizational details were kept, as they are relevant to the credibility of initiatives. The paper provides a new KG resource and a guideline for their usage. We do not claim that the best model for us will perform best in all cases.

    \item \textbf{Regarding the usage of methods implemented in the paper:} Manipulation in the age of generative AI is a concern and a genuine research topic. Our work was done inside an Institute of repute and in association with a global multinational corporate organization. Both of these institutions follow their internal ethics and responsible AI protocols. Manipulation detection/prevention is a separate research topic. Prevention of unethical practices in the current research was done by following the organizational checks and protocols. The resources (dataset and code) generated will be provided on-demand to ensure usage for research purposes only. A declaration statement and research proposal will be mandatory.
\end{enumerate}

\bibliography{custom}

\appendix

\section{Experimental Setup} \label{appendix:exp-setup}
The base LM model is GPT2-medium\citep{radford2019language}, which forms the foundation for generating text. We utilize the GODEL base version 
\footnote{GODEL Model available at \href{https://huggingface.co/microsoft/GODEL-v1_1-base-seq2seq}{Hugging Face}} and Llama3-3B version \footnote{Llama3 Model available at \href{https://huggingface.co/meta-llama/Meta-Llama-3-8B}{Hugging Face}} \footnote{Blenderbot Model available at \href{https://huggingface.co/docs/transformers/en/model_doc/blenderbot}{Hugging Face}}for our experiments. The training was done on four RTX-2080TI-11GB GPUs, with CUDA 11.2, using the \href{https://github.com/huggingface/autotrain-advanced}{Autotrain} package. The experiment took eight days' worth of cumulative runtime.

GPT2, GODEL, Blenderbot and Llama3 were all fine-tuned for 30 epochs with a batch size of 4. The GPT2-rl model was optimized for 438 total steps over a batch size of 128. The intent and other classifier modules were fine-tuned for 15 epochs each for a batch size of 64. 

We compared Open Source models, including Llama3-8B,  \href{https://huggingface.co/docs/transformers/en/model_doc/blenderbot}{Blenderbot V1}, Falcon-7B, MPT-7B, OPT -6.7B, and Alpaca 7-B. Out of these, Llama3 was selected as the best one after internal discussion and our experiences with previous research works.

\section{Human Evaluation Details:}\label{appendix:humanannotationdetails}

\subsection{Data Quality Control}

Eight human participants conducted manual checks and guideline-based scoring to ensure the conversations' internal coherence, content consistency, and naturalness. The team rated the dialogues on a Likert scale from 1 to 5, adhering to predefined guidelines covering grammatical correctness, thematic consistency, language appropriateness, user profile consistency, and clinical sensitivity. The guidelines were as follows:
 \begin{itemize}
   \item Participants were tasked with identifying grammatical errors, subject-verb agreement issues, and improper word usage within the conversations.
   \item They checked for thematic coherence and logical flow, aiming to maintain consistency and avoid abrupt topic shifts or dialogue discontinuity.
   \item Evaluating the appropriateness of language used in the conversations, particularly in terms of natural tone and flow, formality, and cultural sensitivity, was emphasized.
    \item Ensure that user attributes and characteristics remain consistent throughout the conversation to maintain coherence and believability.
   \item Participants were requested to pay special attention to the portrayal of clinical interactions, aiming to enhance politeness and empathy.
\end{itemize}

An inter-evaluator Kappa \citep{mchugh2012interrater} agreement ratio of 80.3\%, 82\%, 84\%  is observed among the experts for internal coherence, content consistency, and naturalness, respectively. After conducting manual checks and receiving feedback from users, the data significantly improved, reflecting a refined level of quality control.

\subsection{Human Evaluation}

We utilized three human annotators (one employed annotators, one non-author Ph.D. scholars, and one masters student) for human evaluation. The inter-annotator agreement between the three human annotators was 83.5\%. We tried with FactScore evaluation method as well but since it was similar to FactTool we eventually decided to choose only one to reduce redundancy in evaluation methods.

\section{Prompts for LLM-based KG creation}\label{appendix:prompts}

We use the following prompts to generate triples from raw text. The list of relations is passed to the prompts from Appendix \ref{appendix:relations}. The relation examples were unnecessary for ChatGPT, but Llama3 required the examples to generate proper triples. Also, the list of sample entities was not necessary for ChatGPT.

\subsection{Llama3}
''' Given the provided relations and sample entities related to environmental sustainability, create a comprehensive set of KG triples by establishing relationships between the entities based on the given relations.

Relations:
    \{Relations\}

Sample entities:

    Rainwater Harvesting
    Permaculture
    Urban Green Spaces
    Regenerative Agriculture
    Ocean Conservation
    Sustainable Transportation
    Wildlife Corridors
    Energy Audits
    Low-impact Development
    Community Gardens
    Green Building Standards
    Forest Stewardship
    Public Transit Infrastructure
    Carbon Sequestration
    Sustainable Forestry
    Energy Star Rating
    Pollution Prevention
    Green Technology
    Food Security
    Land Use Planning

Instructions:

1) Identify and establish relationships with other entities based on the provided relations for each sample entity.
2) Provide KG triples for each relation and entity combination, ensuring logical connections within the environmental sustainability domain.
3) Use the provided sample entities as subjects or objects to construct KG triples.
4) Utilize the relations to create meaningful connections between entities, reflecting various aspects of environmental sustainability such as conservation practices, sustainable technologies, and ecosystem management.
5) Ensure that KG triples are coherent and relevant within environmental sustainability, reflecting real-world relationships and concepts.
6) If there are no more triples, stop.
7) Make sure there are no repeated triples.'''
    
\subsection{ChatGPT}
''' Given the provided relations and sample entities related to environmental sustainability, create a comprehensive set of KG triples by establishing relationships between the entities based on the given relations.

Relations:
    \{Relations\}

Instructions:

    1) Identify and establish relationships with other entities based on the provided relations for each possible entity.
    2) Provide KG triples for each relation and entity combination, ensuring logical connections within the environmental sustainability domain that contain a numerical value, as per the sample triples provided.
    3) Utilize the relations to create meaningful connections between entities, reflecting various aspects of environmental sustainability such as conservation practices, sustainable technologies, and ecosystem management.
    4) Ensure that KG triples are coherent and relevant within environmental sustainability and contain a numerical value, reflecting real-world relationships and concepts.
    5) Make sure there are no repeated triples.
    6) If there are no more triples, stop.'''

\section{Human Curated Fact Statements:}\label{appendix:humancurationdetails}

To properly curate the list of relations, we asked human annotators to generate factual statements to provide environmental sustainability suggestions.

\subsection{Data Quality Control}

Four human participants conducted manual checks and guideline-based scoring to ensure the claims' internal coherence, content consistency, and naturalness. The team rated the dialogues on a Likert scale from 1 to 5, adhering to predefined guidelines covering grammatical correctness, thematic consistency, language appropriateness, user profile consistency, and clinical sensitivity. The guidelines were as follows:
 \begin{itemize}
   \item Participants were tasked with identifying grammatical errors, subject-verb agreement issues, and improper word usage within the conversations.
   \item They checked for thematic coherence and logical flow, aiming to maintain consistency and avoid abrupt topic shifts or dialogue discontinuity.
   \item Evaluating the appropriateness of language used in the conversations, particularly regarding natural tone and flow, formality, and cultural sensitivity, was emphasized.
    \item Ensure that user attributes and characteristics remain consistent throughout the conversation to maintain coherence and believability.
   \item Participants were requested to pay special attention to the portrayal of clinical interactions, aiming to enhance politeness and empathy.
\end{itemize}

An inter-evaluator Kappa \citep{mchugh2012interrater} agreement ratio of 80.3\%, 82\%, 84\%  is observed among the experts for internal coherence, content consistency, and naturalness, respectively. After conducting manual checks and receiving feedback from users, the data significantly improved, reflecting a satisfactory level of quality control.

\section{Support Statements and Relation}\label{appendix:relations}
Given below are examples of support statements and relations that were utilized to create our KG.

\subsection{Support Statements}
Here are a few examples of manually created support statements from which we reverse-engineered a list of relevant relations.
\begin{itemize}
    \item Approximately 8 million metric tons of plastic waste enter the oceans every year, endangering marine life and ecosystems.
    \item Residential energy use accounts for about 20\% of global CO2 emissions.
    \item Deforestation accounts for approximately 11\% of global greenhouse gas emissions.
    \item The transportation sector is responsible for around 14\% of global greenhouse gas emissions.
    \item The average food item in the United States travels about 1,500 miles to reach consumers.
    \item Approximately one-third of all food produced globally goes to waste.
    \item Energy-efficient appliances can reduce energy consumption by up to 50\% compared to standard models.
    \item In 2020, renewable energy sources accounted for 29\% of global electricity generation.
    \item Freshwater scarcity affects over 2 billion people worldwide.
    \item Approximately 1 million animal and plant species are currently threatened with extinction.
\end{itemize}

\subsection{Selected Relations}
Here is a list of the 46 relations we have chosen, along with sample triples that utilize them:

\begin{table*}[ht]
\small
\centering
\begin{tabular}{|l|l|}
\hline
\textbf{Relation} & \textbf{Example} \\
\hline
HasStatistic & (Carbon Emissions, HasStatistic, 4.3 metric tons per capita) \\
HasNumericValue & (Global electricity generation, HasNumericValue, 90\%) \\
HasUnitOfMeasurement & (Water Conservation, HasUnitOfMeasurement, gallons per household per month) \\
HasContext & (Climate Change, HasContext, global temperatures rising) \\
HasSource & (Environmental Report, HasSource, IPCC) \\
HasSubject & (Sustainable Agriculture, HasSubject, soil health) \\
HasAction & (Climate Change Mitigation Strategies, HasAction, Implementing Carbon Pricing) \\
HasAverageValue & (Vehicle Emissions Regulations, HasAverageValue, 100 grams per kilometer) \\
HasMinValue & (Waste Reduction Campaign, HasMinValue, 30\% reduction in waste production) \\
HasMaxValue & (Green Building Standards, HasMaxValue, LEED Platinum certification) \\
HasImpact & (Renewable Energy Project, HasImpact, reducing carbon footprint) \\
HasPercentileValue & (Renewable Energy, HasPercentileValue, 28\% of Electricity Generation) \\
HasTrend & (Climate Action Plan, HasTrend, increasing adoption rates) \\
HasComparison & (Renewable Energy Adoption, HasComparison, surpassing fossil fuel usage) \\
HasImpact & (Electric Vehicle Adoption, HasImpact, reducing air pollution) \\
HasCorrelation & (Wildlife Corridors, HasCorrelation, Biodiversity conservation) \\
Reduces & (Sustainable Transportation, Reduces, carbon emissions) \\
Saves & (Water Conservation Measures, Saves, freshwater resources) \\
Decreases & (Plastic Ban Policy, Decreases, plastic pollution) \\
Increases & (Renewable Energy Incentives, Increases, adoption rates) \\
EfficiencyOf & (Energy Efficiency of Appliances, EfficiencyOf, Energy Star certified products) \\
PercentageOf & (Water Management Facilities, PercentageOf, Over 15\% Total Electricity Consumption) \\
RatioOfFrequencyOf & (Renewable Energy, RatioOfFrequencyOf, Renewable Energy Adoption) \\
RateOf & (Fashion industry, RateOf, 10\% of Carbon Emmisions) \\
VolumeOf & (Volume of Water Used in Agriculture, VolumeOf, 70\%) \\
EmissionOf & (CO2 Emissions, EmissionOf, transportation sector) \\
ConsumptionOf & (Energy Consumption, ConsumptionOf, residential buildings) \\
ImpactOf & (Biodiversity Conservation Efforts, ImpactOf, preserving habitats) \\
BenefitOf & (Renewable Energy Benefits, BenefitOf, reducing dependency on fossil fuels) \\
AdvantageOf & (Sustainable Agriculture, AdvantageOf, improving soil health) \\
DisadvantageOf & (Renewable Energy Adoption, DisadvantageOf, initial high installation costs) \\
RiskOf & (Climate Change, RiskOf, extreme weather events) \\
PreventionOf & (Carbon Capture Technology, PreventionOf, CO2 emissions) \\
ProtectionOf & (Protected Areas, ProtectionOf, endangered species) \\
PreservationOf & (Ecosystem Restoration Projects, PreservationOf, natural habitats) \\
ConservationOf & (Waste Management Programs, ConservationOf, landfill space) \\
RecoveryOf & (Ocean Cleanup Initiatives, RecoveryOf, marine ecosystems) \\
ManagementOf & (Forest Management Practices, ManagementOf, timber resources) \\
RegulationOf & (Environmental Regulations, RegulationOf, industrial emissions) \\
PolicyOf & (Climate Change Policy, PolicyOf, reducing carbon emissions) \\
InitiativeOf & (Sustainability Initiative, InitiativeOf, local government) \\
StrategyOf & (Environmental Strategy, StrategyOf, corporate sustainability) \\
AdaptationOf & (Climate Adaptation Plan, AdaptationOf, changing climate conditions) \\
MitigationOf & (Climate Mitigation Measures, MitigationOf, greenhouse gas emissions) \\
HasCapacity & ("Wind Energy", "HasCapacity", "63 GW") \\
HasPolicyTarget & ("European Union", "HasPolicyTarget", "40\% renewable electricity by 2030") \\
\hline
\end{tabular}
\caption{Examples of relations and their corresponding triples in curated knowledge graphs.}
\label{tab:relations}
\end{table*}

\end{document}